# Stabilizing Gradients for Deep Neural Networks via Efficient SVD Parameterization

Jiong Zhang*, Qi Lei* and Inderjit S. Dhillon*†‡

*Institute for Computational Engineering and Sciences, UT Austin
†Department of Computer Science, UT Austin
‡Amazon/A9
{zhangjiong724@, leiqi@ices.,dhillon@cs.}utexas.edu

March 25, 2018

## Abstract

Vanishing and exploding gradients are two of the main obstacles in training deep neural networks, especially in capturing long range dependencies in recurrent neural networks (RNNs). In this paper, we present an efficient parametrization of the transition matrix of an RNN that allows us to stabilize the gradients that arise in its training. Specifically, we parameterize the transition matrix by its singular value decomposition (SVD), which allows us to explicitly track and control its singular values. We attain efficiency by using tools that are common in numerical linear algebra, namely Householder reflectors for representing the orthogonal matrices that arise in the SVD. By explicitly controlling the singular values, our proposed Spectral-RNN method allows us to easily solve the exploding gradient problem and we observe that it empirically solves the vanishing gradient issue to a large extent. We note that the SVD parameterization can be used for any rectangular weight matrix, hence it can be easily extended to any deep neural network, such as a multi-layer perceptron. Theoretically, we demonstrate that our parameterization does not lose any expressive power, and show how it controls generalization of RNN for the classification task. Our extensive experimental results also demonstrate that the proposed framework converges faster, and has good generalization, especially in capturing long range dependencies, as shown on the synthetic addition and copy tasks, as well as on MNIST and Penn Tree Bank data sets.

## 1 Introduction

Deep neural networks have achieved great success in various fields, including computer vision, speech recognition, natural language processing, etc. Despite their tremendous capacity to fit complex functions, optimizing deep neural networks remains a contemporary challenge. Two main obstacles are vanishing and exploding gradients, that become particularly problematic in Recurrent Neural Networks (RNNs) since the transition matrix is identical at each layer, and any slight change to it is amplified through recurrent layers (4).

Several methods have been proposed to solve the issue, for example, Long Short Term Memory (LSTM) (10) and residual networks (9). Another recently proposed class of methods is designed to enforce orthogonality of the square transition matrices, such as unitary and orthogonal

RNNs (oRNN) (2; 17). However, while these methods solve the exploding gradient problem, they limit the expressivity of the network.

In this paper, we present an efficient parametrization of weight matrices that arise in a deep neural network, thus allowing us to stabilize the gradients that arise in its training, while retaining the desired expressive power of the network. In more detail we make the following contributions:

- We propose a method to parameterize weight matrices through their singular value decomposition (SVD). Inspired by (17), we attain efficiency by using tools that are common in numerical linear algebra, namely Householder reflectors for representing the orthogonal matrices that arise in the SVD. The SVD parametrization allows us to retain the desired expressive power of the network, while enabling us to explicitly track and control singular values.

- We apply our SVD parameterization to recurrent neural networks to exert spectral constraints on the RNN transition matrix. Our proposed Spectral-RNN method enjoys similar space and time complexity as the vanilla RNN. We empirically verify the superiority of Spectral-RNN over RNN/oRNN, in some case even LSTMs, over an exhaustive collection of time series classification tasks and the synthetic addition and copy tasks, especially when the network depth is large.

- Theoretically, we prove that the generalization gap in margin loss of general RNN is bounded by the $t$-th power of the spectral norm of the transition matrices, where $t$ is the depth of layers. Therefore by controlling singular values we can reduce the population risk.

- Our parameterization is general enough to eliminate the gradient vanishing/exploding problem not only in RNNs, but also in various deep networks. We illustrate this by applying SVD parametrization to problems with non-square weight matrices, specifically multi-layer perceptrons (MLPs) and residual networks.

We now present the outline of our paper. In Section 2, we discuss related work, while in Section 3 we introduce our SVD parametrization and demonstrate how it spans the whole parameter space and does not limit expressivity. In Section 4 we propose the Spectral-RNN model that is able to efficiently control and track the singular values of the transition matrices, and we extend our parameterization to non-square weight matrices and apply it to MLPs in Section 5. Section 6 provides the optimization landscape of Spectral-RNN by showing that linear Spectral-RNN has no spurious local minimum. Experimental results on synthetic addition and copy tasks, and on MNIST and Penn Tree Bank data are presented in Section 7. Finally, we present our conclusions and future work in Section 8.

## 2 Related Work

Numerous approaches have been proposed to address the vanishing and exploding gradient problem. Long short-term memory (LSTM) (10) attempts to address the vanishing gradient problem by adding additional memory gates. Residual networks (9) pass the original input directly to the next layer in addition to the original layer output. (18) performs gradient clipping, while (20) apply spectral regularization to the weight matrices. Other approaches include introducing $L_1$ or $L_2$ penalization on successive gradient norm pairs in back propagation (20).

Recently the idea of restricting transition matrices to be orthogonal has drawn some attention. (14) proposed initializing recurrent transition matrices to be identity or orthogonal (IRNN). This strategy shows better performance when compared to vanilla RNN and LSTM. However, there is no guarantee that the transition matrix is close to orthogonal after a few iterations. The unitary RNN (uRNN) algorithm proposed in (2) parameterizes the transition matrix with reflection, diagonal and Fourier transform matrices. By construction, uRNN ensures that the transition matrix is unitary at all times. Although this algorithm performs well on several small tasks, (24) showed that uRNN only covers a subset of possible unitary matrices and thus detracts from the expressive power of RNN. An improvement over uRNN, the orthogonal RNN (oRNN), was proposed by (17). oRNN uses products of Householder reflectors to represent an orthogonal transition matrix, which is rich enough to span the entire space of orthogonal matrices. Meanwhile, (23) empirically demonstrate that the strong constraint of orthogonality limits the model's expressivity, thereby hindering its performance. Therefore, they parameterize the transition matrix by its SVD, $W = U\Sigma V^\top$ (factorized RNN) and restrict $\Sigma$ to be in a range close to 1; however, the orthogonal matrices $U$ and $V$ are updated by geodesic gradient descent using the Cayley transform, thereby resulting in time complexity cubic in the number of hidden nodes which is prohibitive for large scale problems. Motivated by the shortcomings of the above methods, our work in this paper attempts to answer the following questions: *Is there an efficient way to solve the gradient vanishing/exploding problem without hurting expressive power?*

As brought to wide notice in (9), deep neural networks should be able to preserve features that are already good. (8) consolidate this point by showing that deep linear residual networks have no spurious local optima. In our work, we broaden this concept and bring it to the area of recurrent neural networks, showing that each layer is not necessarily near identity, but being close to orthogonality suffices to get a similar result.

Generalization is a major concern in training deep neural networks. (19) and (3) provide a margin-based generalization bound for feedforward neural networks by a spectral Lipschitz constant, namely the product of spectral norm of each layer. We extended the analysis to recurrent neural network and show our scheme of restricting the spectral norm of weight matrices reduces generalization error in the same setting as (19). As supported by the analysis in (6), since our SVD parametrization allows us to develop an efficient way to constrain the weight matrix to be a tight frame (22), we consequently are able to reduce the sensitivity of the network to adversarial examples.

## 3  SVD parameterization

The SVD of the transition matrix $W \in \mathbb{R}^{n \times n}$ of an RNN is given by $W = U\Sigma V^T$, where $\Sigma$ is the diagonal matrix of singular values, and $U, V \in \mathbb{R}^{n \times n}$ are orthogonal matrices, i.e., $U^T U = UU^T = I$ and $V^T V = VV^T = I$ (21). During the training of an RNN, our proposal is to maintain the transition matrix in its SVD form. However, in order to do so efficiently, we need to maintain the orthogonal matrices $U$ and $V$ in compact form, so that they can be easily updated by forward and backward propagation. In order to do so, as in (17), we use a tool that is commonly used in numerical linear algebra, namely Householder reflectors (which, for example, are used in computing the $QR$ decomposition of a matrix).

Given a vector $u \in \mathbb{R}^k, k \leq n$, we define the $n \times n$ Householder reflector $\mathcal{H}_k^n(u)$ to be:

$$\mathcal{H}_k^n(u) = \begin{cases} \begin{pmatrix} I_{n-k} & \\ & I_k - 2\frac{uu^\top}{\|u\|^2} \end{pmatrix} &, \quad u \neq \mathbf{0} \\ I_n &, \quad \text{otherwise.} \end{cases} \quad (1)$$

The Householder reflector is clearly a symmetric matrix, and it can be shown that it is orthogonal, i.e., $H^2 = I$ (11). Further, when $u \neq 0$, it has $n-1$ eigenvalues that are 1, and one eigenvalue which is $-1$ (hence the name that it is a reflector) . In practice, to store a Householder reflector, we only need to store $u \in \mathbb{R}^k$ rather than the full matrix.

Given a series of vectors $\{u_i\}_{i=k}^n$ where $u_k \in \mathbb{R}^k$, we define the map:

$$\begin{aligned}\mathcal{M}_k : \mathbb{R}^k \times ... \times \mathbb{R}^n &\mapsto \mathbb{R}^{n \times n} \\ (u_k, ..., u_n) &\mapsto \mathcal{H}_n(u_n)...\mathcal{H}_k(u_k),\end{aligned} \quad (2)$$

where the right hand side is a product of Householder reflectors, yielding an orthogonal matrix (to make the notation less cumbersome, we remove the superscript from $\mathcal{H}_k^n$ for the rest of this section).

**Theorem 1.** *The image of $\mathcal{M}_1$ is the set of all $n \times n$ orthogonal matrices.*

The proof of Theorem 1 is an easy extension of the Householder QR factorization Theorem, and is presented in Appendix A. Although we cannot express all $n \times n$ matrices with $\mathcal{M}_k$, any $W \in \mathbb{R}^{n \times n}$ can be expressed as the product of two orthogonal matrices $U, V$ and a diagonal matrix $\Sigma$, i.e. by its SVD: $W = U\Sigma V^\top$. Given $\sigma \in \mathbb{R}^n$ and $\{u_i\}_{i=k_1}^n, \{v_i\}_{i=k_2}^n$ with $u_i, v_i \in \mathbb{R}^i$, we finally define our proposed SVD parametrization:

$$\begin{aligned}\mathcal{M}_{k_1,k_2} : \mathbb{R}^{k_1} \times ... \times \mathbb{R}^n \times \mathbb{R}^{k_2} \times ... \times \mathbb{R}^n \times \mathbb{R}^n &\mapsto \mathbb{R}^{n \times n} \\ (u_{k_1}, ..., u_n, v_{k_2}, ..., v_n, \sigma) &\mapsto \mathcal{H}_n(u_n)...\mathcal{H}_{k_1}(u_{k_1}) diag(\sigma) \mathcal{H}_{k_2}(v_{k_2})...\mathcal{H}_n(v_n).\end{aligned}$$

**Theorem 2.** *The image of $\mathcal{M}_{1,1}$ is the set of $n \times n$ real matrices,*
*i.e., $\mathbb{R}^{n \times n} = \mathcal{M}_{1,1}\left(\mathbb{R}^1 \times ... \times \mathbb{R}^n \times \mathbb{R}^1 \times ... \times \mathbb{R}^n \times \mathbb{R}^n\right)$*

The proof of Theorem 2 is based on the singular value decomposition and Theorem 1, and is presented in Appendix A. The astute reader might note that $\mathcal{M}_{1,1}$ seemingly maps an input space of $n^2 + 2n$ dimensions to a space of $n^2$ dimensions; however, since $\mathcal{H}_k^n(u_k)$ is invariant to the norm of $u_k$, the input space also has exactly $n^2$ dimensions. Although Theorems 1 and 2 are simple extensions of well known linear algebra results, they ensure that our parameterization has the ability to represent any matrix and so the full expressive power of the RNN is retained.

**Theorem 3.** *The image of $\mathcal{M}_{k_1,k_2}$ includes the set of all orthogonal $n \times n$ matrices if $k_1 + k_2 \leq n + 2$.*

Theorem 3 indicates that if the total number of reflectors is greater than $n$: $(n - k_1 + 1) + (n - k_2 + 1) \geq n$, then the parameterization covers all orthogonal matrices. Note that when fixing $\sigma = \mathbf{1}$, $\mathcal{M}_{k_1,k_2}(\{u_i\}_{i=k_1}^n, \{v_i\}_{i=k_2}^n, \mathbf{1}) \in \mathbf{O}(n)$, where $\mathbf{O}(n)$ is the set of $n \times n$ orthogonal matrices. Thus when $k_1 + k_2 \leq n + 2$, we have $\mathbf{O}(n) = \mathcal{M}_{k_1,k_2}\left[\mathbb{R}^{k_1} \times ... \times \mathbb{R}^n \times \mathbb{R}^{k_2} \times ... \times \mathbb{R}^n \times \mathbf{1}\right]$.

## 4 Spectral-RNN

In this section, we apply our SVD parameterization to RNNs and describe the resulting Spectral-RNN algorithm in detail. Given a hidden state vector from the previous step $h^{(t-1)} \in \mathbb{R}^n$ and input $x^{(t-1)} \in \mathbb{R}^{n_i}$, RNN computes the next hidden state $h^{(t)}$ and output vector $\hat{y}^{(t)} \in \mathbb{R}^{n_y}$ as:

$$\begin{aligned}h^{(t)} &= \sigma(Wh^{(t-1)} + Mx^{(t-1)} + b), \\ \hat{y}^{(t)} &= Yh^{(t)}.\end{aligned} \quad (3)$$

In Spectral-RNN we parametrize the transition matrix $W \in \mathbb{R}^{n \times n}$ using $m_1 + m_2$ Householder reflectors as:

$$W = \mathcal{M}_{k_1, k_2}(u_{k_1}, ..., u_n, v_{k_2}, ..., v_n, \sigma)$$
$$= \mathcal{H}_n(u_n)...\mathcal{H}_{k_1}(u_{k_1}) diag(\sigma) \mathcal{H}_{k_2}(v_{k_2})...\mathcal{H}_n(v_n)$$

where $k_1 = n - m_1 + 1$, $k_2 = n - m_2 + 1$. This parameterization gives us several advantages over the regular RNN. First, we can select the number of reflectors $m_1$ and $m_2$ to balance expressive power versus time and space complexity. By Theorem 2, the choice $m_1 = m_2 = n$ gives us the same expressive power as vanilla RNN. Notice oRNN could be considered a special case of our parametrization, since when we set $m_1 + m_2 \geq n$ and $\sigma = \mathbf{1}$, we can represent all orthogonal matrices, as proven by Theorem 3. Most importantly, we are able to explicitly control the singular values of the transition matrix. In most cases, we want to constrain the singular values to be within a small interval near 1. The most intuitive method is to clip the singular values that are out of range. Another approach would be to initialize all singular values to 1, and add a penalty term $\|\sigma - 1\|^2$ to the objective function. Here, we have applied another parameterization of $\sigma$ proposed in (23):

$$\sigma_i = 2r(f(\hat{\sigma}_i) - 0.5) + \sigma^*, \ i \in [n] \tag{4}$$

where $f$ is the sigmoid function and $\hat{\sigma}_i$ is updated from $u_i, v_i$ via stochastic gradient descent. The above allows us to constrain $\sigma_i$ to be within $[\sigma^* - r, \sigma^* + r]$. In practice, $\sigma^*$ is usually set to 1 and $r \ll 1$. Note that we are not incurring more computation cost or memory for the parameterization. For regular RNN, the number of parameters is $(n_y + n_i + n + 1)n$, while for Spectral-RNN it is $(n_y + n_i + m_1 + m_2 + 2)n - \frac{m_1^2 + m_2^2 - m_1 - m_2}{2}$. In the extreme case where $m_1 = m_2 = n$, it becomes $(n_y + n_i + n + 3)n$. Later we will show that the computational cost of Spectral-RNN is also of the same order as RNN.

## 4.1 Forward/backward propagation

In forward propagation, we need to iteratively evaluate $h^{(t)}$ from $t = 0$ to $L$ using (10). The only different aspect from a regular RNN in the forward propagation is the computation of $Wh^{(t-1)}$. Note that in Spectral-RNN, $W$ is expressed as product of $m_1 + m_2$ Householder matrices and a diagonal matrix. Thus $Wh^{(t-1)}$ can be computed iteratively using $(m_1 + m_2)$ inner products and vector additions. Denoting $\hat{u}_k = \binom{\mathbf{0}_{n-k}}{u_k}$, we have:

$$\mathcal{H}_k(u_k)h = \left(I_n - \frac{2\hat{u}_k \hat{u}_k^\top}{\hat{u}_k^\top \hat{u}_k}\right)h = h - 2\frac{\hat{u}_k^\top h}{\hat{u}_k^\top \hat{u}_k}\hat{u}_k$$

Thus, the total cost of computing $Wh^{(t-1)}$ is $O((m_1 + m_2)n)$ floating point operations (flops). Detailed analysis can be found in Section 4.2. Let $L(\{u_i\}, \{v_i\}, \sigma, M, Y, b)$ be the loss or objective function, $C^{(t)} = Wh^{(t)}, \hat{\Sigma} = diag(\hat{\sigma})$. Given $\frac{\partial L}{\partial C^{(t)}}$, we define:

$$\frac{\partial L}{\partial u_k^{(t)}} := \left[\frac{\partial C^{(t)}}{\partial u_k^{(t)}}\right]^\top \frac{\partial L}{\partial C^{(t)}}; \quad \frac{\partial L}{\partial v_k^{(t)}} := \left[\frac{\partial C^{(t)}}{\partial v_k^{(t)}}\right]^\top \frac{\partial L}{\partial C^{(t)}};$$

$$\frac{\partial L}{\partial \Sigma^{(t)}} := \left[\frac{\partial C^{(t)}}{\partial \Sigma^{(t)}}\right]^\top \frac{\partial L}{\partial C^{(t)}}; \quad \frac{\partial L}{\partial \hat{\Sigma}^{(t)}} := \left[\frac{\partial \Sigma^{(t)}}{\partial \hat{\Sigma}^{(t)}}\right]^\top \frac{\partial L}{\partial \Sigma^{(t)}};$$

$$\frac{\partial L}{\partial h^{(t-1)}} := \left[\frac{\partial C^{(t)}}{\partial h^{(t-1)}}\right]^\top \frac{\partial L}{\partial C^{(t)}}$$

Back propagation for Spectral-RNN requires $\frac{\partial C^{(t)}}{\partial u_k^{(t)}}, \frac{\partial C^{(t)}}{\partial v_k^{(t)}}, \frac{\partial C^{(t)}}{\partial \hat{\Sigma}^{(t)}}$ and $\frac{\partial C^{(t)}}{\partial h^{(t-1)}}$. These partial gradients can also be computed iteratively by computing the gradient of each Householder matrix at a time. We drop the superscript $(t)$ now for ease of exposition. Given $\hat{h} = \mathcal{H}_k(u_k)h$ and $g = \frac{\partial L}{\partial \hat{h}}$, we have

$$\frac{\partial L}{\partial h} = \left[\frac{\partial \hat{h}}{\partial h}\right]^\top \frac{\partial L}{\partial \hat{h}} = \left(I_n - \frac{2\hat{u}_k \hat{u}_k^\top}{\hat{u}_k^\top \hat{u}_k}\right) g = g - 2\frac{\hat{u}_k^\top g}{\hat{u}_k^\top \hat{u}_k}\hat{u}_k, \tag{5}$$

$$\frac{\partial L}{\partial \hat{u}_k} = \left[\frac{\partial \hat{h}}{\partial \hat{u}_k}\right]^\top \frac{\partial L}{\partial \hat{h}}$$

$$= -2\left(\frac{\hat{u}_k^\top h}{\hat{u}_k^\top \hat{u}_k} I_n + \frac{1}{\hat{u}_k^\top \hat{u}_k} h\hat{u}_k^\top + \frac{\hat{u}_k^\top h}{(\hat{u}_k^\top \hat{u}_k)^2}\hat{u}_k \hat{u}_k^\top\right) g$$

$$= -2\frac{\hat{u}_k^\top h}{\hat{u}_k^\top \hat{u}_k} g - 2\frac{\hat{u}_k^\top g}{\hat{u}_k^\top \hat{u}_k} h + 4\frac{\hat{u}_k^\top h}{\hat{u}_k^\top \hat{u}_k}\frac{\hat{u}_k^\top g}{\hat{u}_k^\top \hat{u}_k}\hat{u}_k. \tag{6}$$

Details of forward and backward propagation can be found in Appendix B.

### 4.2 Complexity Analysis

Table 1 gives the time complexity of various algorithms. $Hprod$ and $Hgrad$ are defined in Algorithm 2 3 (see Appendix B). Algorithm 2 needs $6k$ flops, while Algorithm 3 uses $(3n + 10k)$ flops. Since $\|u_k\|^2$ only needs to be computed once per iteration, we can further decrease the flops to $4k$ and $(3n + 8k)$. Also, in back propagation we can reuse $\alpha$ in forward propagation to save $2k$ flops. The efficiency of training Spectral-RNN can be improved by adopting the level 3 BLAS, blocked

|  | flops |
|---|---|
| $Hprod(h, u_k)$ | $4k$ |
| $Hgrad(h, u_k, g)$ | $3n + 6k$ |
| Spectral-RNN-Local FP$(n, m_1, m_2)$ | $4n(m_1 + m_2) - 2m_1^2 - 2m_2^2 + O(n)$ |
| Spectral-RNN-Local BP$(n, m_1, m_2)$ | $6n(m_1 + m_2) - 1.5m_1^2 - 1.5m_2^2 + O(n)$ |
| oRNN-Local FP$(n, m)$ | $4nm - m^2 + O(n)$ |
| oRNN-Local BP$(n, m)$ | $7nm - 2m^2 + O(n)$ |

Table 1: Time complexity across algorithms

Householder QR algorithm (1) to exploit GPU computing power.

# 5 Extending SVD Parameterization to General Weight Matrices

In this section, we extend the parameterization to non-square matrices and use Multi-Layer Perceptrons(MLP) as an example to illustrate its application to general deep networks. For any weight matrix $W \in \mathbb{R}^{m \times n}$(without loss of generality $m \leq n$), its reduced SVD can be written as:

$$W = U(\Sigma|0)(V_L|V_R)^\top = U\Sigma V_L^\top, \tag{7}$$

where $U \in \mathbb{R}^{m \times m}$, $\Sigma \in diag(\mathbb{R}^m)$, $V_L \in \mathbb{R}^{n \times m}$. There exist $u_n, ..., u_{k_1}$ and $v_n, ..., v_{k_2}$ s.t. $U = \mathcal{H}_m^m(u_m)...\mathcal{H}_{k_1}^m(u_{k_1})$, $V = \mathcal{H}_n^n(v_n)...\mathcal{H}_{k_2}^n(v_{k_2})$, where $k_1 \in [m], k_2 \in [n]$. Thus we can extend the SVD parameterization for any non-square matrix:

$$\begin{aligned}
\mathcal{M}_{k_1,k_2}^{m,n} : &\mathbb{R}^{k_1} \times ... \times \mathbb{R}^m \times \mathbb{R}^{k_2} \times ... \times \mathbb{R}^n \times \mathbb{R}^{\min(m,n)} \\
&\mapsto \mathbb{R}^{m \times n} \\
&(u_{k_1}, ..., u_m, v_{k_2}, ..., v_n, \sigma) \\
&\mapsto \mathcal{H}_m^m(u_m) \cdots \mathcal{H}_{k_1}^m(u_{k_1})\hat{\Sigma}\mathcal{H}_{k_2}^n(v_{k_2}) \cdots \mathcal{H}_n^n(v_n). 
\end{aligned} \tag{8}$$

where $\hat{\Sigma} = (diag(\sigma)|0)$ if $m < n$ and $(diag(\sigma)|0)^\top$ otherwise. Next we show that we only need $2\min(m,n)$ reflectors (rather than $m+n$) to parametrize any $m \times n$ matrix. By the definition of $\mathcal{H}_k^n$, we have the following lemma:

**Lemma 1.** *Given $\{v_i\}_{i=1}^n$, define $V^{(k)} = \mathcal{H}_n^n(v_n)...\mathcal{H}_k^n(v_k)$ for $k \in [n]$. We have:*

$$V_{*,i}^{(k_1)} = V_{*,i}^{(k_2)}, \forall k_1, k_2 \in [n], i \leq \min(n-k_1, n-k_2).$$

Here $V_{*,i}$ indicates the $i$th column of matrix $V$. According to Lemma 1, we only need at most first $m$ Householder vectors to express $V_L$, which results in the following Theorem:

**Theorem 4.** *If $m \leq n$, the image of $\mathcal{M}_{1,n-m+1}^{m,n}$ is the set of all $m \times n$ matrices; else the image of $\mathcal{M}_{n-m+1,1}^{m,n}$ is the set of all $m \times n$ matrices.*

Similarly if we constrain $u_i, v_i$ to have unit length, the input space dimensions of $\mathcal{M}_{1,n-m+1}^{m,n}$ and $\mathcal{M}_{m-n+1,1}^{m,n}$ are both $mn$, which matches the output dimension. Thus we extend Theorem 2 to the non-square case, which enables us to apply SVD parameterization to not only the RNN transition matrix, but also to general weight matrices in various deep learning models. For example, the Multilayer perceptron (MLP) model is a class of feedforward neural network with fully connected layers:

$$h^{(t)} = f(W^{(t-1)}h^{(t-1)} + b^{(t-1)}) \tag{9}$$

Here $h^{(t)} \in \mathbb{R}^{n_t}$, $h^{(t-1)} \in \mathbb{R}^{n_{t-1}}$ and $W^{(t)} \in \mathbb{R}^{n_t \times n_{t-1}}$. Applying SVD parameterization to $W^{(t)}$ say $n_t < n_{t-1}$, we have:

$$\begin{aligned}
W^{(t)} =& \mathcal{H}_{n_t}^{n_t}(u_{n_t})...\mathcal{H}_1^{n_t}(u_1)\Sigma \\
& \cdot \mathcal{H}_{n_{t-1}-n_t+1}^{n_{t-1}}(v_{n_{t-1}-n_t+1})...\mathcal{H}_{n_{t-1}}^{n_{t-1}}(v_{n_{t-1}}).
\end{aligned}$$

We can use the same forward/backward propagation algorithm as described in Algorithm 1. Besides RNN and MLP, our SVD parameterization also applies to more advanced frameworks, such as Residual networks and LSTM, which we will not describe in detail here.

# 6 Theoretical Analysis

Since we can control and upper bound the singular values of the transition matrix in Spectral-RNN, we can clearly eliminate the exploding gradient problem. In this section, we provide the first generalization analysis for RNN, and prove that by upper bounding the singular values of transition matrix, our Spectral-RNN ensures good generalization.

## 6.1 Generalization bound for RNN

To study the generalization of general recurrent neural network, we simplify the network by absorbing the bias term $b$ in $M$ and consider:

$$h^{(t)} = \sigma(Wh^{(t-1)} + Mx^{(t-1)}), h^{(0)} = 0, \tag{10}$$

$$\hat{y}^{(t)} = Yh^{(t)}. \tag{11}$$

For simplicity, here we assume $x^{(t)}, \hat{y}^{(t)}, h^{(t)} \in \mathbb{R}^n$, $W, M, Y \in \mathbb{R}^{n \times n}$, and input for each layer is bounded by $B: \|x^{(t)}\| \leq B, t = 0, 1, \cdots$. For a classification task, we consider the following *Margin Loss* defined in (19):

**Definition 1.** *For any distribution $\mathcal{D}$ and margin $\gamma > 0$, we define the expected margin loss as follows:*

$$L_\gamma(f_w) = \mathbb{P}_{(x,y) \sim \mathcal{D}} \left[ f_w(x)[y] \leq \gamma + \max_{j \neq y} f_w(x)[j] \right],$$

where $f_w(x)[y]$ is the probability of predicting $y$ given input $x$ with weight $w$. We use $\hat{L}_\gamma$ to represent the empirical margin loss.

**Theorem 5.** *For any $B, t, n > 0$, let $\hat{y}_w : \mathbb{R}^{n \times t} \to \mathbb{R}^n$ be a $t$-layer recurrent neural network with ReLU activations, where input $\{x^{(0)}, \cdots x^{(t-1)}\}$ satisfies $\|x^{(i)}\| \leq B, \forall i = 0, 1, \cdots t-1$. Then, for any $\delta, \gamma > 0$, with probability $\geq 1 - \delta$ over a training set of size $m$, for any $w$, we have:*

$$L_0(f_w) \leq \hat{L}_\gamma(f_w) + \mathcal{O}\left(\sqrt{\frac{B(w) + \ln \frac{tm}{\delta}}{m}}\right),$$

where $B(w) = B^2 t^2 n \ln(n) \max\{\|W\|_2^{2t-2}, 1\} \|M\|_2^2 \|Y\|_2^2 \cdot \left(t^2 \|W\|_F^2 + \frac{\|M\|_F^2}{\|M\|_2^2} + \frac{\|Y\|_F^2}{\|Y\|_2^2}\right) / \gamma^2$.

From Theorem 5, we can see that $W$ plays a huge role since generalization gap grows exponentially with $\|W\|$, i.e. the largest singular value of $W$. Meanwhile our proposed Spectral-RNN, which bounds the singular radius of $W$ in $[1-r, 1+r]$, ensures good generalization:

**Corollary 1.** *With the update rule in (4), Spectral-RNN has generalization gap bounded by $\mathcal{O}(\sqrt{\frac{B^2 t^4 n^2 \ln(n)(1+r)^{2t} \|M\|_2^2 \|Y\|_2^2 / \gamma^2 + \ln \frac{tm}{\delta}}{m}})$ with probability $\geq 1 - \delta$.*

The proof of Theorem 5 is presented in the Appendix A, which uses the PAC-Bayes strategy similar as it is in (19): a combination of the PAC-Bayes margin analysis (Lemma 3 from (16)) and the following perturbation analysis of the neural network in Lemma 2 we derived.

**Lemma 2.** *Write $w = vec(\{W, Y, M\})$, and perturbation $u = vec(\{\delta W, \delta Y, \delta M\})$ such that $\|\delta W\| \leq \frac{1}{t}\|W\|$, $\|\delta Y\| \leq \frac{1}{t}\|Y\|, \|\delta M\| \leq \frac{1}{t}\|M\|$. For a t-layered recurrent neural network, the perturbations in the activation is bounded by:*

$$\|h_{w+u}^{(t)}(x) - h_w^{(t)}(x)\| \leq Bte\|\delta M\| + B(t+2)\|\delta W\|\|M\|\|W\|^{t-2}.$$

*While the perturbation in the output satisfies:*

$$|\hat{y}_{w+u}^{(t)}(x) - \hat{y}_w^{(t)}(x)| \leq tB \max\{\|W\|^{t-1}, 1\} \cdot (\|Y\|\|\delta W\|\|M\|te + \|Y\|\|\delta M\|e + \|\delta Y\|\|M\|)$$

While the above analysis lends further credence to our observed experimental results, we leave it to future work to extend the analysis for more general tasks other than the classification problems.

## 7 Experimental Results

In this section, we provide empirical evidence that shows the advantages of SVD parameterization in both RNNs and MLPs. For RNN models, we compare our Spectral-RNN algorithm with (vanilla) RNN, IRNN (14), oRNN (17) and LSTM (10). The transition matrix in IRNN is initialized to be orthogonal while other matrices are initialized by sampling from a Gaussian distribution. For MLP models, we implemented vanilla MLP, Residual Network (ResNet) (9) and applied SVD parameterization on both of them. We used a residual block of two layers in ResNet. In most cases *leaky_Relu* is used as activation function except for LSTM. To train these models, we applied Adam optimizer with stochastic gradient descent (13). These models are implemented with Tensorflow (7).[1][2] Other than the experiments reported in this section, we provide UCR time series classification and multi-label learning results in Appendix C.

### 7.1 Addition and Copy tasks

We tested RNN models on the Addition and Copy tasks with the same settings as (2).

**Addition task:** The Addition task requires the network to remember two marked numbers in a long sequence and add them. Each input data includes two sequences: top sequence whose values are sampled uniformly from [0, 1] and bottom sequence which is a binary sequence with only two 1's. The network is asked to output the sum of the two values. From the empirical results in Figure 1, we can see that when the network is not deep (number of layers $L$=30 in (a)(d)), every model outperforms the baseline of 0.167 (always output 1 regardless of the input). Also, the first layer gradients do not vanish for all models. However, on longer sequences ($L$=100 in (b)(e)), IRNN fails and LSTM converges much slower than Spectral-RNN and oRNN. If we further increase the sequence length ($L$=300 in (c)(f)), only Spectral-RNN and oRNN are able to beat the baseline within a reasonable number of iterations. We can also observe that the first layer gradient of oRNN/Spectral-RNN does not vanish regardless of the depth, while IRNN/LSTM's gradients vanish as $L$ becomes lager.

---

[1] We thank Mhammedi for providing their code for oRNN (17)
[2] Our code is available at https://github.com/zhangjiong724/spectral-RNN

Figure 1: RNN models on the addition task with $L$ layers and hidden dimension of $n_h$. The top plots show the test MSE, while the bottom plots show the magnitude of the gradient at each corresponding step.

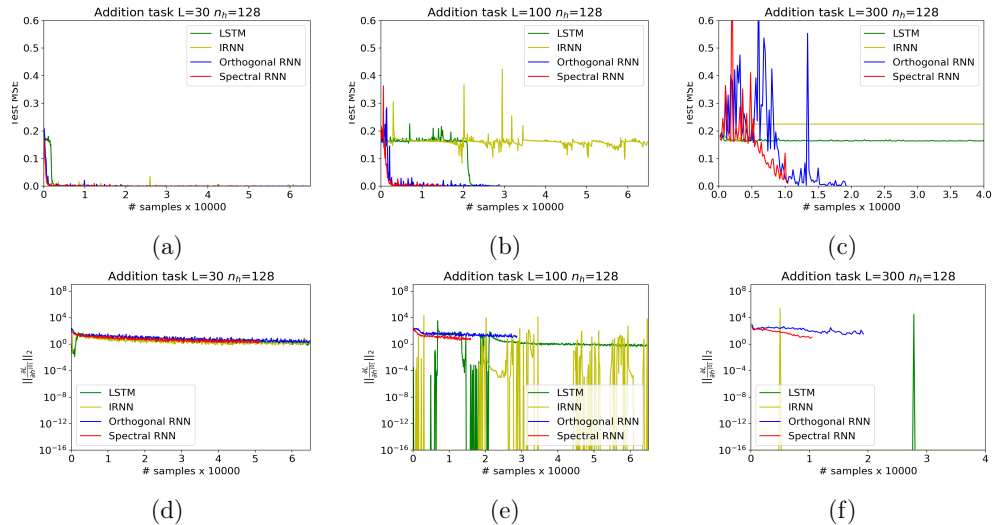

Figure 2: RNN models on the Copy task with time lag $T$ and hidden dimension $n_h$.

**Copy task:** Let $A = \{a_i\}_{i=0}^{9}$ be the alphabet. The input data sequence $x \in A^{T+20}$ where $T$ is the time lag. $x_{1:10}$ are sampled uniformly from $\{a_i\}_{i=0}^{7}$ and $x_{T+10}$ is set to $a_9$. The rest of $x_i$ are set to $a_8$. The network is asked to output $x_{1:10}$ after seeing $a_9$. That is to copy $x_{1:10}$ from the beginning to the end with time lag $T$.

A baseline strategy is to predict $a_8$ for $T+10$ entries and randomly sample from $\{a_i\}_{i=1}^{7}$ for the last 10 digits. From the empirical results in Figure 2, Spectral-RNN consistently outperforms all other models. IRNN and LSTM models are not able to beat the baseline when the time lag is large. In fact, the test MSE for RNN/LSTM is very close to the baseline (memoryless strategy) indicating that they do not memorize any useful information throughout the larger time lag.

### 7.2 pixel-MNIST and permute-MNIST

In this experiment, we compare different models on the MNIST image dataset. The dataset was split into a training set of 60000 instances and a test set of 10000 instances. The $28 \times 28$ MNIST pixels are flattened into a vector and then traversed by the RNN models. Table 2 shows test accuracy across multiple models. Spectral-RNN reaches the highest 97.6% accuracy on pixel-MNIST with only 128 hidden dimensions and 6k parameters.

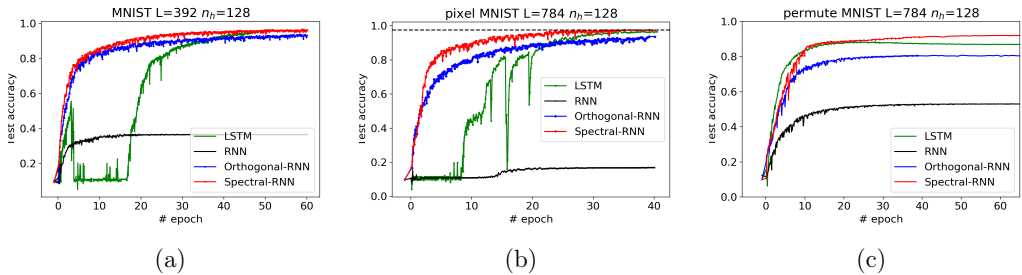

|     |     |     |
| :-: | :-: | :-: |
| (a) | (b) | (c) |

Figure 3: RNN models on pixel-MNIST and permute-MNIST. Spectral-RNN constantly yields the highest test accuracy.

| Models | Hidden dimension | # parameters | Test accuracy |
| --- | :---: | :---: | :---: |
| Spectral-RNN | $128(m_1, m_2 = 16)$ | $\approx 6k$ | **97.7** |
| oRNN (17) | $256(m = 32)$ | $\approx 11k$ | 97.2 |
| RNN (23) | 128 | $\approx 35k$ | 94.1 |
| uRNN (2) | 512 | $\approx 16k$ | 95.1 |
| RC uRNN (24) | 512 | $\approx 16k$ | 97.5 |
| FC uRNN (24) | 116 | $\approx 16k$ | 92.8 |
| factorized RNN (23) | 128 | $\approx 32k$ | 94.6 |
| LSTM (23) | 128 | $\approx 64k$ | 97.3 |

Table 2: Results for pixel MNIST across multiple algorithms

Figure 3(a)(b) plots the test accuracy on networks with 392 and 784 temporal steps respectively. We also tested models on the permuted-MNIST dataset, where we apply a fixed random permutation to the pixels before training. We performed a grid search over several learning rates $\rho = \{0.1, 0.01, 0.001, 0.0001\}$, decay rate $\alpha = \{0.9, 0.8, 0.5\}$ and batch size $B = \{64, 128, 256, 512\}$. The reported results are the best one among them. Figure 3(c) shows the test accuracy on permuted MNIST dataset. Also we explored the effect of different spectral constraints and explicitly tracked the spectral margin ($\max_i |\sigma_i - 1|$) of the transition matrix. Figure 4 shows the spectral margin of different RNN models. Although IRNN has small spectral margin at first few iterations, it quickly deviates from being orthogonal. Figure 4 shows the magnitude of the first layer gradient $\|\frac{\partial L}{\partial h^{(0)}}\|_2$ during training. RNN suffers from vanishing gradient at first several epochs, LSTM's gradient explode after several epochs while oRNN and Spectral-RNN have much more stable gradients.

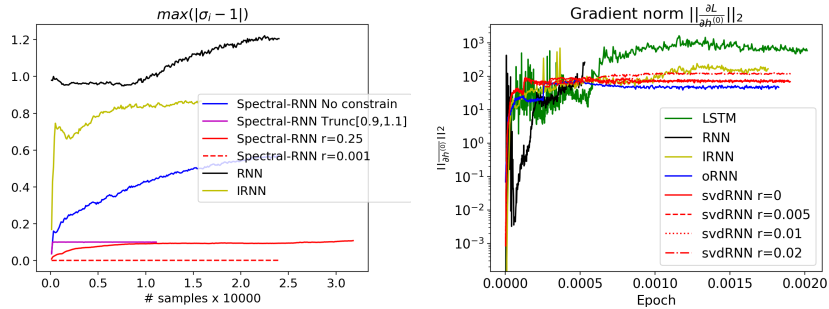

Figure 4: $\sigma$ deviation and gradient magnitude

For the MLP models, each instance is flattened to a vector of length 784 and fed to the input layer. After the input layer there are 30-100 layers with hidden dimension 128 (Figure 5). On a shallow network, Spectral-MLP and Spectral-ResNet achieve similar performance as ResNet while MLP's convergence is slower. However, when the network is deeper, both MLP and ResNet start to fail. MLP is not able to function with $L > 35$ and ResNet with $L > 70$. On the other hand, the SVD based methods are resilient to increasing depth and thus achieve higher precision.

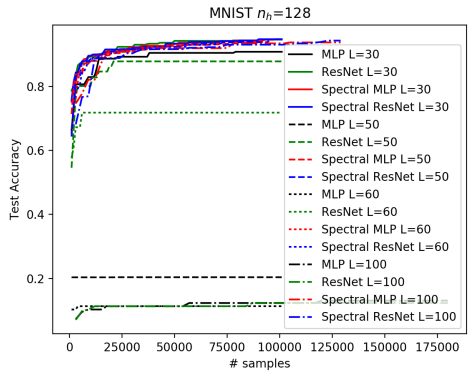

Figure 5: MLP models on MNIST with $L$ layers and $n_h$ hidden dimension. Spectral-based methods are resilient to increasing depth.

### 7.3 Penn Tree Bank dataset

We tested different models on Penn Tree Bank (PTB) (15) dataset for word-level prediction tasks. The dataset contains 929k training words, 73k validation words, and 82k test words with 10k vocabulary. We trained 1- and 2-layered RNN models on word sequences of length 300. We adopted the successive mini-batches method (25), that use the final hidden state of the previous mini-batch as the initial state of the next one. We use initial learning rate of 0.1 and decay by factor of 0.8 at each epoch, and 80% dropout is applied on 2-layered models.

| Models($n_l$,$n_h$) | # parameters | Train perplexity | Test perplexity |
|---|---|---|---|
| RNN(1,128) | $\approx 16k$ | 68.1 | 144.7 |
| LSTM(1,128) | $\approx 64k$ | 69.1 | 130.7 |
| Spectral-RNN(1,512) | $\approx 31k$ | 65.4 | 130.2 |
| RNN(2,128) | $\approx 32k$ | 62.6 | 142.5 |
| LSTM(2,128) | $\approx 128k$ | 26.1 | 122.7 |
| Spectral-RNN(2,512) | $\approx 63k$ | 36.0 | 121.3 |

Table 3: Penn Tree Bank word level prediction

As seen in Table 3, Spectral-RNN achieves better performance than LSTM with about half the number of parameters. Note that 2-layered Spectal-RNN achieves lower text perplexity with higher training perplexity, which shows its generalization ability.

# 8 Conclusions

In this paper, we have proposed an efficient SVD parametrization of weight matrices in deep neural networks, which allows us to explicitly track and control their singular values. This parameterization does not restrict the network's expressive power, while simultaneously allowing fast forward as well as backward propagation. The method is easy to implement and has the same time and space complexity as compared to original methods like RNN and MLP. The ability to control singular values helps in avoiding the gradient vanishing and exploding problems, and as we have empirically shown, gives good performance. However, further experimentation is required to fully understand the influence of using different number of reflectors in our SVD parameterization. Also, the underlying structures of the image of $\mathcal{M}_{k_1,k_2}$ when $k_1, k_2 \neq 1$ is a subject worth investigating.

# References

[1] Robert Andrew and Nicholas Dingle. Implementing QR factorization updating algorithms on GPUs. *Parallel Computing*, 40(7):161–172, 2014.

[2] Martin Arjovsky, Amar Shah, and Yoshua Bengio. Unitary evolution recurrent neural networks. In *International Conference on Machine Learning*, pages 1120–1128, 2016.

[3] Peter Bartlett, Dylan J Foster, and Matus Telgarsky. Spectrally-normalized margin bounds for neural networks. *arXiv preprint arXiv:1706.08498*, 2017.

[4] Yoshua Bengio, Patrice Simard, and Paolo Frasconi. Learning long-term dependencies with gradient descent is difficult. *IEEE transactions on neural networks*, 5(2):157–166, 1994.

[5] Yanping Chen, Eamonn Keogh, Bing Hu, Nurjahan Begum, Anthony Bagnall, Abdullah Mueen, and Gustavo Batista. The UCR time series classification archive, July 2015. www.cs.ucr.edu/~eamonn/time_series_data/.

[6] Moustapha Cisse, Piotr Bojanowski, Edouard Grave, Yann Dauphin, and Nicolas Usunier. Parseval networks: Improving robustness to adversarial examples. In *International Conference on Machine Learning*, pages 854–863, 2017.

[7] Martín Abadi et. al. TensorFlow: Large-scale machine learning on heterogeneous systems, 2015. Software available from tensorflow.org.

[8] Moritz Hardt and Tengyu Ma. Identity matters in deep learning. *arXiv preprint arXiv:1611.04231*, 2016.

[9] Kaiming He, Xiangyu Zhang, Shaoqing Ren, and Jian Sun. Deep residual learning for image recognition. In *Proceedings of the IEEE conference on computer vision and pattern recognition*, pages 770–778, 2016.

[10] Sepp Hochreiter and Jürgen Schmidhuber. Long short-term memory. *Neural computation*, 9(8):1735–1780, 1997.

[11] Alston S Householder. Unitary triangularization of a nonsymmetric matrix. *Journal of the ACM (JACM)*, 5(4):339–342, 1958.

[12] Michael Hüsken and Peter Stagge. Recurrent neural networks for time series classification. *Neurocomputing*, 50:223–235, 2003.

[13] Diederik Kingma and Jimmy Ba. Adam: A method for stochastic optimization. *arXiv preprint arXiv:1412.6980*, 2014.

[14] Quoc V Le, Navdeep Jaitly, and Geoffrey E Hinton. A simple way to initialize recurrent networks of rectified linear units. *arXiv preprint arXiv:1504.00941*, 2015.

[15] Mitchell P Marcus, Mary Ann Marcinkiewicz, and Beatrice Santorini. Building a large annotated corpus of english: The Penn Treebank. *Computational linguistics*, 19(2):313–330, 1993.

[16] David McAllester. Simplified PAC-Bayesian margin bounds. In *Learning theory and Kernel machines*, pages 203–215. Springer, 2003.

[17] Zakaria Mhammedi, Andrew Hellicar, Ashfaqur Rahman, and James Bailey. Efficient orthogonal parametrisation of recurrent neural networks using Householder reflections. In *International Conference on Machine Learning*, pages 2401–2409, 2017.

[18] Tomáš Mikolov. Statistical language models based on neural networks. *Presentation at Google, Mountain View, 2nd April*, 2012.

[19] Behnam Neyshabur, Srinadh Bhojanapalli, David McAllester, and Nathan Srebro. A PAC-Bayesian approach to spectrally-normalized margin bounds for neural networks. *arXiv preprint arXiv:1707.09564*, 2017.

[20] Razvan Pascanu, Tomas Mikolov, and Yoshua Bengio. On the difficulty of training recurrent neural networks. In *International Conference on Machine Learning*, pages 1310–1318, 2013.

[21] Lloyd N Trefethen and David Bau III. *Numerical linear algebra*, volume 50. SIAM, 1997.

[22] Joel A Tropp, Inderjit S Dhillon, Robert W Heath, and Thomas Strohmer. Designing structured tight frames via an alternating projection method. *IEEE Transactions on information theory*, 51(1):188–209, 2005.

[23] Eugene Vorontsov, Chiheb Trabelsi, Samuel Kadoury, and Chris Pal. On orthogonality and learning recurrent networks with long term dependencies. In *International Conference on Machine Learning*, pages 3570–3578, 2017.

[24] Scott Wisdom, Thomas Powers, John Hershey, Jonathan Le Roux, and Les Atlas. Full-capacity unitary recurrent neural networks. In *Advances in Neural Information Processing Systems*, pages 4880–4888, 2016.

[25] Wojciech Zaremba, Ilya Sutskever, and Oriol Vinyals. Recurrent neural network regularization. *arXiv preprint arXiv:1409.2329*, 2014.

# Appendix A   Proofs

## A.1   Proof of Proposition 1

**Proposition 1.** *(Householder QR factorization)* Let $B \in \mathbb{R}^{n \times n}$. Then there exists an upper triangular matrix $R$ with positive diagonal elements, and vectors $\{u_i\}_{i=1}^n$ with $u_i \in \mathbb{R}^i$, such that $B = \mathcal{H}_n^n(u_n)...\mathcal{H}_1^n(u_1)R$. (Note that we allow $u_i = 0$, in which case, $H_i^n(u_i) = I_n$ as in (1))

*Proof of Proposition 1.* For $n = 1$, note that $\mathcal{H}_1^1(u_1) = \pm 1$. By setting $u_1 = 0$ if $B_{1,1} > 0$ and $u_1 \neq 0$ otherwise, we have the factorization desired.

Assume that the result holds for $n = k$, then for $n = k+1$ set $u_{k+1} = B_1 - \|B_1\|e_1$. Here $B_1$ is the first column of $B$ and $e_1 = (1, 0, ..., 0)^\top$. Thus we have

$$\mathcal{H}_{k+1}^{k+1}(u_{k+1})B = \begin{pmatrix} \|B_1\| & \hat{B}_{1,2:k+1} \\ 0 & \hat{B} \end{pmatrix},$$

where $\hat{B} \in \mathbb{R}^{k \times k}$. Note that $\mathcal{H}_{k+1}^{k+1}(u_{k+1}) = I_{k+1}$ when $u_{k+1} = 0$ and the above still holds. By assumption we have $\hat{B} = \mathcal{H}_k^k(u_k)...\mathcal{H}_1^k(u_1)\hat{R}$. Notice that $\mathcal{H}_i^{k+1}(u_i) = \begin{pmatrix} 1 & \\ & \mathcal{H}_i^k(u_i) \end{pmatrix}$, so we have that

$$\mathcal{H}_1^{k+1}(u_1)...\mathcal{H}_k^{k+1}(u_k)\mathcal{H}_{k+1}^{k+1}(u_{k+1})B = \begin{pmatrix} \|B_1\| & \tilde{B}_{1,2:k+1} \\ 0 & \hat{R} \end{pmatrix} = R$$

is an upper triangular matrix with positive diagonal elements. Thus the result holds for any $n$ by the theory of mathematical induction. □

## A.2   Proof of Theorem 1

*Proof.* Observe that the image of $\mathcal{M}_1$ is a subset of $\mathbf{O}(n)$, and we now show that the converse is also true. Given $A \in \mathbf{O}(n)$, by Proposition 1, there exists an upper triangular matrix $R$ with positive diagonal elements, and an orthogonal matrix $Q$ expressed as $Q = \mathcal{H}_n^n(u_n)...\mathcal{H}_1^n(u_1)$ for some set of Householder vectors $\{u_i\}_{i=1}^n$, such that $A = QR$. Since $A$ is orthogonal, we have $A^\top A = AA^\top = I_n$, thus:

$$A^\top A = R^\top Q^\top QR = R^\top R = I_n; \quad Q^\top AA^\top Q = Q^\top QRR^\top Q^\top Q = RR^\top = I_n$$

Thus $R$ is orthogonal and upper triangular matrix with positive diagonal elements. So $R = I_n$ and $A = Q = \mathcal{H}_n^n(u_n)...\mathcal{H}_1^n(u_1)$. □

## A.3   Proof of Theorem 2

*Proof.* It is easy to see that the image of $\mathcal{M}_{1,1}$ is a subset of $\mathbb{R}^{n \times n}$. For any $W \in \mathbb{R}^{n \times n}$, we have its SVD, $W = U\Sigma V^\top$, where $\Sigma = diag(\sigma)$. By Theorem 1, for any orthogonal matrix $U, V \in \mathbb{R}^{n \times n}$, there exists $\{u_i\}_{i=1}^n \{v_i\}_{i=1}^n$ such that $U = \mathcal{M}_1(u_1, ..., u_n)$ and $V = \mathcal{M}_1(v_1, ..., v_n)$, then we have:

$$W = \mathcal{H}_n^n(u_n)...\mathcal{H}_1^n(u_1)\Sigma\mathcal{H}_1^n(v_1)...\mathcal{H}_n^n(v_n)$$
$$= \mathcal{M}_{1,1}(u_1, ..., u_n, v_1, ..., v_n, \sigma)$$

□

## A.4   Proof of Theorem 3

*Proof.* Let $A \in \mathbb{R}^{n \times n}$ be an orthogonal matrix. By Theorem 1, there exist $\{a_i\}_{i=1}^n$, such that $A = \mathcal{M}_1(a_1, ..., a_n)$. Since $A^\top$ is also orthogonal, for the same reason, there exist $\{b_i\}_{i=1}^n$, such that $A^\top = \mathcal{M}_1(b_1, ..., b_n)$. Thus we have:

$$A = \mathcal{H}_n(a_n)...\mathcal{H}_1(a_1) = \mathcal{H}_1(b_1)...\mathcal{H}_n(b_n)$$

Observe that one of $k_2 \geq k_1 - 1$ and $k_1 \geq k_2 - 1$ must be true. If $k_2 \geq k_1 - 1$, set

$$u_k = a_k, k = n, n-1, ..., k_1,$$
$$v_{k_2+k_1-k-1} = a_k, k = k_1 - 1, ..., 1, \quad (12)$$
$$v_t = \mathbf{0}, t = k_2 + k_1 - 2, ..., n,$$

and then we have:

$$\begin{aligned}
\mathcal{M}_{k_1,k_2}(u_{k_1},...,u_n,v_{k_2},...,v_n,\mathbf{1}) &= \mathcal{H}_n(u_n)...\mathcal{H}_{k_1}(u_{k_1})I_n\mathcal{H}_{k_2}(v_{k_2})...\mathcal{H}_n(v_n) \\
&= \mathcal{H}_n(a_n)...\mathcal{H}_{k_1}(a_{k_1})I_n\mathcal{H}_{k_1-1}(a_{k_1-1})...\mathcal{H}_1(a_1) \\
&= A
\end{aligned} \quad (13)$$

Else, assign:

$$v_k = b_k, k = n, n-1, ..., k_2,$$
$$u_{k_2+k_1-k-1} = b_k, k = k_2 - 1, ..., 1, \quad (14)$$
$$u_t = \mathbf{0}, t = k_2 + k_1 - 2, ..., n,$$

and then we have:

$$\begin{aligned}
\mathcal{M}_{k_1,k_2}(u_{k_1},...,u_n,v_{k_2},...,v_n,\mathbf{1}) &= \mathcal{H}_1(b_1)...\mathcal{H}_{k_2-1}(b_{k_2-1})I_n\mathcal{H}_{k_2}(b_{k_2})...\mathcal{H}_n(b_n) \\
&= A
\end{aligned} \quad (15)$$

$\square$

## A.5 Proof of Theorem 4

*Proof.* It is easy to see that the image of $\mathcal{M}_{*,*}^{m,n}$ is a subset of $\mathbb{R}^{m \times n}$. For any $W \in \mathbb{R}^{m \times n}$, we have its SVD, $W = U\Sigma V^\top$, where $\Sigma$ is an $m \times n$ diagonal matrix. By Theorem 1, for any orthogonal matrix $U \in \mathbb{R}^{m \times m}, V \in \mathbb{R}^{n \times n}$, there exists $\{u_i\}_{i=1}^m \{v_i\}_{i=1}^n$ such that $U = \mathcal{H}_m^m(u_m)...\mathcal{H}_1^m(u_1)$ and $V = \mathcal{H}_n^n(v_n)...\mathcal{H}_1^n(v_1)$. By Lemma 1, if $m < n$ we have:

$$\begin{aligned}
W &= \mathcal{H}_n^m(u_n)...\mathcal{H}_1^m(u_1)\Sigma\mathcal{H}_1^n(v_1)...\mathcal{H}_n^n(v_n) \\
&= \mathcal{H}_n^m(u_n)...\mathcal{H}_1^m(u_1)\Sigma\mathcal{H}_{n-m+1}^n(v_{n-m+1})...\mathcal{H}_n^n(v_n).
\end{aligned}$$

Similarly, for $n < m$, we have:

$$\begin{aligned}
W &= \mathcal{H}_n^m(u_n)...\mathcal{H}_1^m(u_1)\Sigma\mathcal{H}_1^n(v_1)...\mathcal{H}_n^n(v_n) \\
&= \mathcal{H}_n^m(u_n)...\mathcal{H}_{m-n+1}^m(u_{m-n+1})\Sigma\mathcal{H}_1^n(v_1)...\mathcal{H}_n^n(v_n).
\end{aligned}$$

$\square$

## A.6 Proof of Theorem 5

To get a generalization bound for RNN, we need to use the following lemma from (16). Recall that $L_0$ as the expected error with 0 margin, and write $\hat{L}_\gamma$ as the empirical error with $\gamma$ margin,

**Lemma 3.** *(16) Let $f_w(x) : \mathcal{X} \to \mathbb{R}^k$ be any predictor (not necessarily a neural network) with parameters $w$, and $P$ be any distribution on the parameters that is independent of the training data. Then, for any $\gamma, \delta > 0$, with probability $\geq 1 - \delta$ over the training set of size $m$, for any $w$, and any random perturbation $u$ s.t. $\mathbb{P}_u[\max_{x \in \mathcal{X}} |f_{w+u}(x) - f_w(x)|_\infty < \frac{\gamma}{4}] \geq \frac{1}{2}$, we have:*

$$L_0(f_w) \leq \hat{L}_\gamma(f_w) + 4\sqrt{\frac{KL(w+u||P) + \ln \frac{6m}{\delta}}{m-1}}$$

In order for $u$ to satisfy the probability property in Lemma 3, we study the change in output with respect to perturbation $u$.

**Lemma 2.** *Write $w = vec(\{W, Y, M\})$, and perturbation $u = vec(\{\delta W, \delta Y, \delta M\})$ such that $\|\delta W\| \leq \frac{1}{t}\|W\|$, $\|\delta Y\| \leq \frac{1}{t}\|Y\|, \|\delta M\| \leq \frac{1}{t}\|M\|$. For a $t$-layered recurrent neural network, the perturbation in the activation $h$ is bounded by:*

$$\|h^{(t)}_{w+u}(x) - h^{(t)}_w(x)\| \leq Bte\|\delta M\| + B(t+2)\|\delta W\|\|M\|\|W\|^{t-2},$$

*and the perturbation in the output satisfies:*

$$|\hat{y}^{(t)}_{w+u}(x) - \hat{y}^{(t)}_w(x)| \leq \quad tB\max\{\|W\|^{t-1}, 1\}(\|Y\|\|\delta W\|\|M\|te + \|Y\|\|\delta M\|e + \|\delta Y\|\|M\|)$$

Proof of Lemma 2 requires to bound the norm of $h$:

**Lemma 4.** $\|h^{(i)}_w\|_2 \leq B\|M\|i\max\{\|W\|^{i-1}, 1\}, i = 1, 2, \cdots, t$

*Proof of Lemma 4.*

$$\|h^{(i)}_w\| \leq \|W\|_2 \|h^{(i-1)}_w\| + \|M\|\|x^{(i-1)}\|$$
$$\leq \cdots$$
$$\leq \|M\| \sum_{j=0}^{i-1} \|W\|_2^{i-1-j} \|x^{(j)}\|$$
$$\leq B\|M\|i\max\{\|W\|^{i-1}, 1\}$$

□

Now we could prove Lemma 2.

*Proof of Lemma 2.* We denote $\Delta_i = \|h^{(i)}_{w+u}(x) - h^{(t)}_w(x)\|$ for short, and prove by induction that $\Delta_i \leq Bi(1+\frac{1}{t})^{i-1}(\|\delta W\|\|M\|t + \|\delta M\|)\max\{\|W\|^{i-1}, 1\}$ for $i \leq t$.

Firstly, $\Delta_0 = 0$ satisfies the inequality. Suppose $i-1$ satisfies the assumption, then for $i \leq t$,

$$\Delta_i = \|\sigma((W + \delta W)h^{(i-1)}_{w+u} - Wh^{(i-1)}_w + (M + \delta M)x^{(i-1)} - Mx^{(i-1)})\|$$
$$= \|\sigma((W + \delta W)(h^{(i-1)}_{w+u} - h^{(i-1)}_w) + \delta W h^{(i-1)}_w + \delta M x^{(i-1)})\|$$
$$\leq (1 + \frac{1}{t})\|W\|\Delta_{i-1} + \|\delta W\|\|h^{(i-1)}_w\| + \|\delta M\|B$$

Then by induction and Lemma 4, we have:

$$\Delta_i \leq (1 + \frac{1}{t})\|W\|B(i-1)(1 + \frac{1}{t})^{i-2}(\|\delta W\|\|M\|t + \|\delta M\|)\max\{\|W\|^{i-2}, 1\}$$
$$+ B(i-1)\|\delta W\|\|M\|\max\{\|W\|^{i-2}, 1\} + B\|\delta M\|$$
$$\leq B\|\delta W\|\|M\|it(1 + \frac{1}{t})^{i-1}\max\{\|W\|^{i-1}, 1\} + B\|\delta M\|i(1 + \frac{1}{t})^{i-1}\max\{\|W\|^{i-1}, 1\}$$
$$= Bi(1 + \frac{1}{t})^{i-1}(t\|\delta W\|\|M\| + \|\delta M\|)\max\{\|W\|^{i-1}, 1\}$$

Therefore for $i = t$, $\Delta_t \leq Bte(t\|M\|\|\delta W\| + \|\delta M\|) \max\{\|W\|^{t-1}, 1\}$. For the perturbation of output $\hat{y}$,

$$\|\hat{y}_{w+u}^{(t)}(x) - \hat{y}_w^{(t)}(x)\|$$
$$= \|(Y + \delta Y)h_{w+u}^{(t)} - Yh_w^{(t)}\| = \|(Y + \delta Y)\Delta_t + \delta Y h_w^{(t)}\|$$
$$\leq \|Y\|(1 + \frac{1}{t})Bt(1 + \frac{1}{t})^{t-1}(t\|\delta W\|\|M\| + \|\delta M\|)\max\{\|W\|^{t-1}, 1\}$$
$$+ \|\delta Y\| tB\|M\| \max\{\|W\|^{t-1}, 1\}$$
$$= tB \max\{\|W\|^{t-1}, 1\}(\|Y\|\|\delta W\|\|M\| te + \|Y\|\|\delta M\| e + \|\delta Y\|\|M\|)$$

□

Finally we are able to prove Theorem 5:

*Proof of Theorem 5.* Choose the distribution of the prior $P = \mathcal{N}(0, \sigma^2 I_{3n \times 3n})$ and consider the random perturbation $u = \text{vec}(\{\delta W, \delta Y, \delta M\})$ with the same zero mean Gaussian distribution, where $\sigma$ will be assigned later. Without loss of generality, we assume $\|Y\|_2, \|M\|_2 \geq 1$.

Since $W, Y, M \sim \mathcal{N}(0, \sigma^2 I_{n \times n})$, we get the following bound for the spectral norm of $\delta W, \delta U, \delta M$:

$$\mathbb{P}_{\delta W \sim \mathcal{N}(0, \sigma^2 I)}[\|\delta W\|_2 > t] \leq 2ne^{-t^2/2n\sigma^2}$$

$$\mathbb{P}_{\delta Y \sim \mathcal{N}(0, \sigma^2 I)}[\|\delta Y\|_2 > t] \leq 2ne^{-t^2/2n\sigma^2}$$

$$\mathbb{P}_{\delta M \sim \mathcal{N}(0, \sigma^2 I)}[\|\delta M\|_2 > t] \leq 2ne^{-t^2/2n\sigma^2}$$

Therefore with probability $\geq \frac{1}{2}$, $\|\delta W\|_2, \|\delta Y\|_2, \|\delta M\|_2 \leq \sigma\sqrt{2n \ln(12n)}$.

Let $\beta = \max\{\|W\|_2^{t-1}, 1\}\|Y\|_2\|M\|_2$. To set up the prior, we define a pre-determined $\tilde{\beta}$ such that $|\tilde{\beta} - \beta| \leq \frac{1}{t}\beta$, and hence $1/e\beta^{t-1} \leq \tilde{\beta}^{t-1} \leq e\beta^{t-1}$.

Plugging into Lemma 2 we have with probability at least $\frac{1}{2}$,

$$\max_{x \in X_{B,n}} |\hat{y}_{w+u}(x) - \hat{y}_w(x)|$$
$$\leq tB \max\{\|W\|^{t-1}, 1\}(\|Y\|\|\delta W\|\|M\| te + \|Y\|\|\delta M\| e + \|\delta Y\|\|M\|)$$
$$\leq tB\sqrt{2n \ln(12n)}\tilde{\beta}\sigma(te + e + 1)$$
$$\leq \frac{\gamma}{4},$$

where we choose $\sigma = \frac{\gamma}{12\sqrt{2n \ln(12n)}Bt(te+e+1)\tilde{\beta}}$. Therefore now the perturbation $u$ satisfies assumptions in Lemma 3.

We next compute the KL-divergence of distributions for P and $u$ for the sake of Lemma 3.

$$KL(w + u \| P) \leq \frac{\|w\|^2}{2\sigma^2}$$
$$\leq \mathcal{O}\left(\frac{B^2 t^2 n \ln(n) \max\{\|W\|^{2t-2}, 1\}\|M\|_2^2\|Y\|_2^2}{\gamma^2}(t^2\|W\|_F^2 + \frac{\|M\|_F^2}{\|M\|_2^2} + \frac{\|Y\|_F^2}{\|Y\|_2^2})\right)$$

□

Hence, with probability $\geq 1 - \delta$ and for all $w$ such that, $|\beta - \tilde{\beta}| \leq \frac{1}{t}\beta$, we have:

$$L_0(\hat{y}_w) \leq \hat{L}_\gamma(\hat{y}_w) + \mathcal{O}(\sqrt{\frac{B(w) + ln\frac{m}{\delta}}{m}}), \tag{16}$$

where $B(w) = \frac{B^2 t^2 n ln(n) \max\{\|W\|^{2t-2}, 1\}\|M\|_2^2\|Y\|_2^2}{\gamma^2}(t^2\|W\|_F^2 + \frac{\|M\|_F^2}{\|M\|_2^2} + \frac{\|Y\|_F^2}{\|Y\|_2^2})$.

Since $\tilde{\beta}$ should be independent of the learned models. We finally take a union bound over different choices of the parameter. We will choose discrete set of $\tilde{\beta}$ such that they cover the real $W, M, Y$ that satisfies $|\beta - \tilde{\beta}| \le \frac{1}{t}\beta$.

Notice $\hat{y}^{(t)} = Yh_w^{(t)}$, therefore if $\beta \le \frac{\gamma}{2Bt}$, $\hat{y} < \frac{\gamma}{2}$ and $\hat{L}_\gamma = 1$ and the bound is satisfied trivially. Meanwhile, when $\beta \ge \frac{\gamma\sqrt{m}}{2Bt}$, then the second term of (16) $\ge 1$ and it also holds trivially. Therefore, we only need to consider $\tilde{\beta}$ such that $\tilde{\beta} \in [\frac{\gamma}{2Bt}, \frac{\gamma\sqrt{m}}{2Bt}]$, and therefore the size of the cover we need to consider is only $tm^{\frac{1}{2t}}$. We take a union bound over the $\tilde{\beta}$ and use (16) to complete the proof.

## Appendix B  Details of Forward and Backward Propagation Algorithms

---

**Algorithm 1** Local forward/backward propagation

**Input**: $h^{(t-1)}, \frac{\partial L}{\partial C^{(t)}}, U = (u_n|...|u_{n-m_1+1})$,
$\Sigma, V = (v_n|...|v_{n-m_2+1})$
**Output**: $C^{(t)} = Wh^{(t-1)}, \frac{\partial L}{\partial U}, \frac{\partial L}{\partial V}, \frac{\partial L}{\partial \hat{\sigma}}, \frac{\partial L}{\partial h^{(t-1)}}$
// Begin forward propagation
$h_{n+1}^{(v)} \leftarrow h^{(t-1)}$
**for** $k = n, n-1, ..., n-m_2+1$ **do**
  $h_k^{(v)} \leftarrow Hprod(h_{k+1}^{(v)}, v_k)$   // Compute $\hat{V}^\top h$
**end for**
$h_{k_1-1}^{(u)} \leftarrow \Sigma h_{k_2}^{(v)}$   // Compute $\Sigma \hat{V}^\top h$
**for** $k = n-m_1+1, ..., n$ **do**
  $h_k^{(u)} \leftarrow Hprod(h_{k-1}^{(u)}, u_k)$   // Compute $\hat{U}\Sigma\hat{V}^\top h$
**end for**
$C^{(t)} \leftarrow h_n^{(u)}$
//Begin backward propagation
$g \leftarrow \frac{\partial L}{\partial C^{(t)}}$
**for** $k = n, n-1, ..., n-m_1+1$ **do**
  $g, G_{*,n-k+1}^{(u)} \leftarrow Hgrad(h_k^{(u)}, u_k, g)$   // Compute $\frac{\partial L}{\partial u_k}$
**end for**
$\bar{\Sigma} \leftarrow diag(g \circ h_{k_2}^{(v)}), g \leftarrow \Sigma g$   // Compute $\frac{\partial L}{\partial \Sigma}$
$g^{(\hat{\sigma})} \leftarrow \frac{\partial diag(\Sigma)}{\partial \hat{\sigma}} \circ diag(\bar{\Sigma})$   // Compute $\frac{\partial L}{\partial \hat{\sigma}}$
**for** $k = n-m_2+1, ..., n$ **do**
  $g, G_{*,n-k+1}^{(v)} \leftarrow Hgrad(h_{k+1}^{(u)}, v_k, g)$   // Compute $\frac{\partial L}{\partial v_k}$
**end for**
$\frac{\partial L}{\partial U} \leftarrow G^{(u)}, \frac{\partial L}{\partial V} \leftarrow G^{(v)}, \frac{\partial L}{\partial \hat{\sigma}} \leftarrow g^{(\hat{\sigma})}, \frac{\partial L}{\partial h^{(t-1)}} \leftarrow g$

---

**Algorithm 2**
$\hat{h} = Hprod(h, u_k)$

**Input**: $h, u_k$
**Output**: $\hat{h} = \mathcal{H}_k(u_k)h$
// Compute $\hat{h} = (I - \frac{2u_k u_k^\top}{u_k^\top u_k})h$
$\alpha \leftarrow \frac{2}{\|u_k\|^2} u_k^\top h$
$\hat{h} \leftarrow h - \alpha u_k$

---

**Algorithm 3**
$\bar{h}, \bar{u}_k = Hgrad(h, u_k, g)$

**Input**: $h, u_k, g = \frac{\partial L}{\partial C}$ where $C = \mathcal{H}_k(u_k)h$
**Output**: $\bar{h} = \frac{\partial L}{\partial h}, \bar{u}_k = \frac{\partial L}{\partial u_k}$
$\alpha = \frac{2}{\|u_k\|^2} u_k^\top h$
$\beta = \frac{2}{\|u_k\|^2} u_k^\top g$
$\bar{h} \leftarrow g - \beta u_k$
$\bar{u}_k \leftarrow -\alpha g - \beta h + \alpha\beta u_k$

---

## Appendix C  More Experimental Details

### C.1  Time Series Classification

In this experiment, we focus on the time series classification problem, where time series are fed into RNN sequentially, which then tries to predict the right class upon receiving the sequence end (12). The dataset we choose is the largest public collection of class-labeled time-series with widely varying length, namely, the UCR time-series collection from (5). We use the training and testing sets directly from the UCR time series archive http://www.cs.ucr.edu/~eamonn/time_series_data/, and randomly choose 20% of the training set as validation data. We provide the statistical descriptions of the datasets and experimental results in Table 4.

In all experiments, we used hidden dimension $n_h = 32$, and chose total number of reflectors for oRNN and Spectral-RNN to be $m = 16$ (for Spectral-RNN $m_1 = m_2 = 8$). We choose proper depth $t$ as well as input size $n_i$. Given sequence length $L$, since $tn_i = L$, we choose $n_i$ to be the maximum divisor of $L$ that satisfies $depth \leq \sqrt{L}$. To have a fair comparison of how the proposed principle itself influences the training procedure, we did not use dropout in any of these models. As illustrated in the optimization process in Figure 6, this resulted in some overfitting (see (a) CBF), but on the other hand it shows that Spectral-RNN is able to prevent overfitting. This supports our claim that since generalization is bounded by the spectral norm of the weights (3), Spectral-RNN will potentially generalize better than other schemes. This phenomenon is more drastic when the depth is large (e.g. ArrowHead(251 layers) and FaceAll(131 layers)), since regular RNN, and even LSTM, have no control over the spectral norms. Also note that there are substantially fewer parameters in oRNN and Spectral-RNN as compared to LSTM.

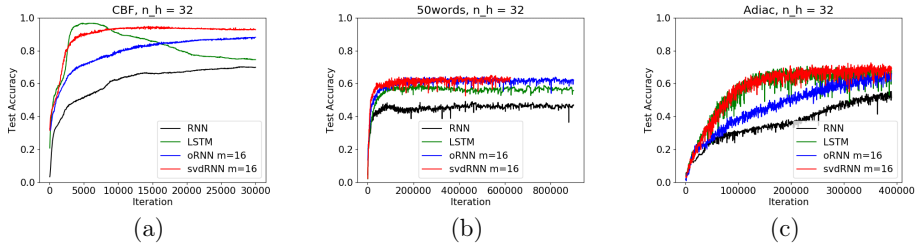

Figure 6: Performance comparisons of the RNN based models on three UCR datasets.

| Datasets | Data Descriptions | | | Depth | RNN | | LSTM | | oRNN | | Spectral-RNN | |
| --- | --- | --- | --- | --- | --- | --- | --- | --- | --- | --- | --- | --- |
| | training/testing size | length | #class | | $acc$ | ($n_{param}$) | $acc$ | ($n_{param}$) | $acc$ | ($n_{param}$) | $acc$ | ($n_{param}$) |
| 50words | 450  455 | 270 | 50 | 27 | 0.492 | (3058) | 0.598 | (7218) | 0.642 | (2426) | **0.651** | (2850) |
| Adiac | 390  391 | 176 | 37 | 16 | 0.552 | (2694) | 0.706 | (6950) | 0.668 | (2062) | **0.726** | (2486) |
| ArrowHead | 36  175 | 251 | 3 | 251 | 0.509 | (1219) | 0.537 | (4515) | 0.669 | (587) | **0.800** | (1011) |
| Beef | 30  30 | 470 | 5 | 47 | 0.600 | (1606) | 0.700 | (5766) | **0.733** | (974) | **0.733** | (1398) |
| BeetleFly | 20  20 | 512 | 2 | 32 | **0.950** | (1699) | 0.850 | (6435) | 0.900 | (1067) | **0.950** | (1491) |
| CBF | 30  900 | 128 | 3 | 16 | 0.702 | (1476) | **0.967** | (5444) | 0.881 | (844) | 0.948 | (1268) |
| Coffee | 28  28 | 286 | 2 | 22 | **1.000** | (1570) | **1.000** | (6018) | **1.000** | (938) | **1.000** | (1362) |
| Cricket X | 390  390 | 300 | 12 | 20 | 0.310 | (1997) | 0.456 | (6637) | 0.495 | (1365) | **0.500** | (1789) |
| DistalPhalanxOutlineCorrect | 276  600 | 80 | 2 | 10 | 0.790 | (1410) | 0.798 | (5378) | 0.830 | (778) | **0.840** | (1202) |
| DistalPhalanxTW | 154  399 | 80 | 6 | 10 | **0.815** | (1641) | 0.795 | (5609) | 0.807 | (1009) | **0.815** | (1433) |
| ECG200 | 100  100 | 96 | 2 | 12 | **0.640** | (1410) | **0.640** | (5378) | **0.640** | (778) | **0.640** | (1202) |
| ECG5000 | 500  4500 | 140 | 5 | 14 | 0.941 | (1606) | 0.936 | (5766) | 0.940 | (974) | **0.945** | (1398) |
| ECGFiveDays | 23  861 | 136 | 2 | 17 | 0.947 | (1443) | 0.790 | (5411) | **0.976** | (811) | 0.948 | (1235) |
| FaceAll | 560  1690 | 131 | 14 | 131 | 0.549 | (1615) | 0.455 | (4911) | **0.714** | (983) | **0.714** | (1407) |
| FaceFour | 24  88 | 350 | 4 | 25 | 0.625 | (1701) | 0.477 | (6245) | 0.511 | (1069) | **0.716** | (1493) |
| FacesUCR | 200  2050 | 131 | 14 | 131 | 0.449 | (1615) | 0.629 | (4911) | 0.710 | (983) | **0.727** | (1407) |
| Gun Point | 50  150 | 150 | 2 | 15 | 0.947 | (1507) | 0.920 | (5667) | 0.953 | (875) | **0.960** | (1299) |
| InsectWingbeatSound | 220  1980 | 256 | 11 | 16 | 0.534 | (1996) | 0.515 | (6732) | **0.598** | (1364) | 0.586 | (1788) |
| ItalyPowerDemand | 67  1029 | 24 | 2 | 6 | 0.970 | (1315) | 0.969 | (4899) | 0.972 | (683) | **0.973** | (1107) |
| Lighting2 | 60  61 | 637 | 2 | 49 | **0.541** | (1570) | **0.541** | (6018) | **0.541** | (938) | **0.541** | (1362) |
| MiddlePhalanxOutlineCorrect | 291  600 | 80 | 2 | 10 | 0.793 | (1410) | 0.783 | (5378) | 0.712 | (778) | **0.820** | (1202) |

Table 4: Test accuracy (number of parameters) on UCR datasets. For each dataset, we present the testing accuracy when reaching the smallest validation error. The highest precision is in bold, and lowest two are colored gray.

### C.1.1 Multi-label Learning

Multi-label learning try to tag a datapoint with the most relevant subset of labels from a large label set. Thus each label $y$ is a high dimensional sparse vector and binary cross entropy is used as the loss function while output activation function is selected as sigmoid.

| Dataset | Feature dimension | Label dimension | $N_{train}$ | $N_{test}$ |
|---------|-------------------|-----------------|-------------|------------|
| Bibtex | 1836 | 159 | 4880 | 2515 |
| Mediamill | 120 | 101 | 30993 | 12914 |

Table 5: Multi-label learning datasets(Prabhu et. al 2014).

From the empirical results in Figure 7, Spectral-ResNet consistently outperforms all other models, especially when network is deep.

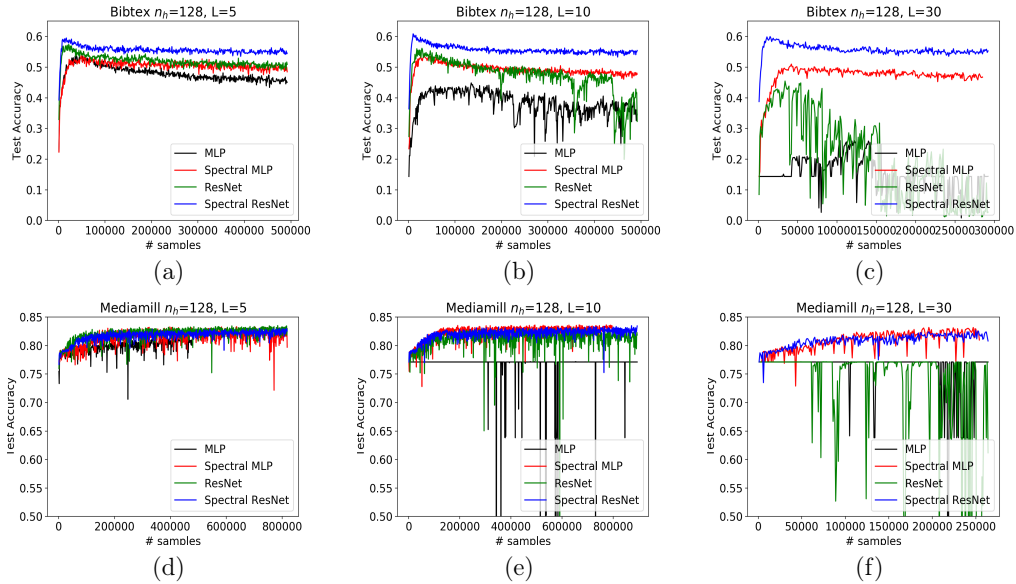

Figure 7: MLP models on multi-label learning with $L$ layers and 128 hidden dimension. Dropout rate of 0.1 is used.